\documentclass[letterpaper]{article} 
\usepackage{aaai24}  
\usepackage{times}  
\usepackage{helvet}  
\usepackage{courier}  
\usepackage[hyphens]{url}  
\usepackage{graphicx} 
\usepackage[caption=false]{subfig}
\usepackage{booktabs}
\usepackage{natbib}  
\usepackage{caption} 
\usepackage{algorithm}
\usepackage{algorithmic}

\usepackage{newfloat}
\usepackage{listings}
\DeclareCaptionStyle{ruled}{labelfont=normalfont,labelsep=colon,strut=off} 
\floatstyle{ruled}
\newfloat{listing}{tb}{lst}{}
\floatname{listing}{Listing}
\title{Plug and Play with Prompts:\\
A Prompt Tuning Approach for Controlling Text Generation}
\author{
    Rohan Deepak Ajwani\textsuperscript{\rm 1,\rm 2},
    Zining Zhu\textsuperscript{\rm 1,\rm 2,\rm 3},
    Jonathan Rose\textsuperscript{\rm 1},
    Frank Rudzicz\textsuperscript{\rm 1,\rm 2,\rm 4}
}
\affiliations{
    \textsuperscript{\rm 1}University of Toronto\\
    \textsuperscript{\rm 2}Vector Institute for Artificial Intelligence\\
    \textsuperscript{\rm 3}Stevens Institute of Technology\\
    \textsuperscript{\rm 4}Dalhousie University\\
    \{rohan, zining\}@cs.toronto.edu, jonathan.rose@utoronto.ca, frank@dal.ca
    
}

\usepackage{bibentry}

\begin{document}

\maketitle

\begin{abstract}
Transformer-based Large Language Models (LLMs) have shown exceptional language generation capabilities in response to text-based prompts. However, controlling the direction of generation via textual prompts has been challenging, especially with smaller models.
In this work, we explore the use of Prompt Tuning to achieve controlled language generation. Generated text is steered using prompt embeddings, which are trained using a small language model, used as a discriminator.
Moreover,  we demonstrate that these prompt embeddings can be trained with a very small dataset, with as low as a few hundred training examples. Our method thus offers a data and parameter efficient solution towards controlling language model outputs.
We carry out extensive evaluation on four datasets: SST-5 and Yelp (sentiment analysis), GYAFC (formality) and JIGSAW (toxic language). Finally, we demonstrate the efficacy of our method towards mitigating harmful, toxic, and biased text generated by language models.

\end{abstract}

\section{Introduction}
With the advent of the transformer architecture \citep{vaswani2017attention}, large language models (LLMs) trained on large amounts of text have become almost ubiquitous for various language processing tasks. Despite their ability to generate grammatically correct and fluent texts, real-world applications often necessitate more specific control over text features beyond fluency, which might not be achieved by allowing LLMs to generate text freely.  

Another challenge posed by unconstrained text generation is the inadvertent production of harmful text. With the abundance of misleading, harmful and defamatory text available online, these harmful biases find their way into the training data of language models. Thus, it has become increasingly important to be able to control the generation of text, and steer language models away from generating biased and toxic outputs.

Controlled generation aims at generating text containing a specific attribute or set of attributes. The usual method to steer the features of LLMs is fine-tuning, which requires training almost all the model's parameters. This method is both data-intensive and computationally inefficient. Moreover, fine-tuning can distort the original pre-trained features of the language model \citep{kumar2022finetuning}, affecting the controllability of the language model. Recent advancements in parameter efficient methods such as prompt tuning \citep{lester-etal-2021-power}, Low-Rank Adaptation (LoRA) \citep{hu2021lora} and adapter tuning \citep{pmlr-v97-houlsby19a} have shown to reduce the trainable parameters by a large factor. However, it is still unknown whether they can achieve effective control while maintaining the fluency in problem scenarios with small amounts of data.

We introduce a novel method, Plug and Play with Prompts (PPP) to address the challenges posed by low-resource controlled generation scenarios.
In our work, we aim to achieve soft control, which controls the generation direction (e.g., the sentiment), as opposed to hard control, which aims to satisfy specific conditions, for example, ensuring that specific words are included in the generated text \citep{pascual-etal-2021-plug-play}. 
PPP utilizes the gradients of an external discriminator model to tune the prompt parameters into control commands that steer the language model generation to achieve soft control. Additionally, we apply the categorical cross entropy loss (CCE) in novel way towards preserving the fluency of the text generated by the prompts.
Our contributions can be summarized as follows:
\begin{enumerate}
    \item We propose PPP, a parameter- and data-efficient method for controllable generation.
    \item With quantitative and qualitative evaluation, we show that PPP can steer the generation style using as few as several hundred training samples. We run ablation studies to show the impacts of hyperparameters.
    \item We additionally show that PPP can generalize to a large out-of-domain training data setting.
\end{enumerate}

\section{Related Work} \label{related_work}
\paragraph{Neural Text Generation:} Neural text generation aims at making neural network based language models generate human-like text. These models are typically trained using maximum likelihood estimation, where the generation task is treated as a multi-label classification problem, and the language model learns to maximize the likelihood of the observed data. This probability model is parameterized autoregressively, where the probability of a generated token is conditioned on the sequence preceding it. The language model used could be based on recursion, including RNNs, LSTMs \citep{hochreiter1997long} and GRUs \citep{cho-etal-2014-learning}, which encode the input sequence as a hidden vector to generate a probability distribution for the next token; or the transformer architecture \citep{vaswani2017attention}, which use attention \citep{bahdanau2016neural} allowing them to focus on more relevant parts of the input, thus increasing context awareness, improving performance for longer sequences, and eliminating the need to process inputs sequentially. During inference, the input is provided to the language model, and the probability distribution generated by the language model is used to obtain the next token. Various decoding algorithms can be used to this end, including greedy decoding, beam search, top-k sampling \citep{fan-etal-2018-hierarchical, holtzman-etal-2018-learning, Radford2019LanguageMA}, and top-p (nucleus) sampling \citep{holtzman2020curious}.

\paragraph{Controlled Generation:} Several methods have been studied towards controlling various attributes of generated text. \citet{ziegler2020finetuning} use reinforcement learning to fine-tune pre-trained language models, whereas \citet{yu2017seqgan} use generative adversarial networks for controlling generation. \citet{keskar2019ctrl} trained a large model (1.63 billion parameters) to generate text conditioned on control codes, which allow users to specify the style, domain, topic, task, and various other attributes while generating text. However, this method requires a large amount of training data to train billions of parameters, making it very expensive. \citet{Dathathri2020Plug} use the gradients of a discriminator to dynamically update the activations of a decoder model at each generation step. \citet{pascual-etal-2021-plug-play} introduced Keyword2Text, which used cosine similarities between a keyword and the vocabulary to update the log-probabilities at each decoding step, and increased the probability of the keyword appearing in the generated text. \citet{yang-klein-2021-fudge} introduced FUDGE, which uses a classifier to modify the output probabilities of a language model while generating text to increase the probability of an attribute appearing in future text. GeDi \citep{krause-etal-2021-gedi-generative} uses class conditioned language models as discriminators to guide the generation. \citet{chan2022cocon} added an additional block (CoCon) within a pre-trained decoder transformer architecture to control the generation of text based on the given content. \citet{chen-etal-2018-stable} and \citet{gu-etal-2017-learning, gu-etal-2017-trainable} explored methods to steer decoding by manipulating the hidden states of the decoder networks.

\paragraph{Prompting:}
Prompting involves providing a pre-trained language model with an instruction (zero-shot) or one or more examples (one/few-shot) demonstrating the task. \citet{brown2020language} showed that prompt design in zero, one, and few-shot settings is very effective at instructing a frozen GPT-3 model to perform a given task. The initial efforts in prompt-tuning were concentrated on discrete selection of prompt template tokens \citep{jiang-etal-2020-know, shin-etal-2020-autoprompt, schick-schutze-2021-exploiting, schick-schutze-2021-just}. More recent works \citep{lester-etal-2021-power, li2021prefixtuning} instead used continuous prompts which were trained using backpropagation while freezing all or most of the model parameters. \citet{hambardzumyan-etal-2021-warp} employ adversarial reprogramming \citep{elsayed2018adversarial} towards learning task specific word embeddings. \citet{chowdhury2022novelty} used prompt tuning towards generating paraphrases. In our work, we build on the work of \citet{lester-etal-2021-power} and attach prompts only to the inputs as embeddings to control the style of the generated text.

\paragraph{Text Style Transfer:} Text style transfer is a similar area that studies the generation of text conditioned on both the content and style of the input text. \citet{pmlr-v70-hu17e} and \citet{shen2017style} disentangle style from content to generate text with a specified style. \citet{shen2020educating} use denoising adversarial autoencoders to map text to a latent space, and change the style in this latent space.

\section{Problem Formulation}
We define controlled generation as a constrained version of autoregressive generation. Given an input sequence of $m$ tokens $x_{1:m} = (x_1, x_2, ..., x_m)$, we want the language model to generate $n$ output tokens $y_{1:n} = (y_1, y_2, ..., y_n)$. The input is an incomplete phrase, and the output is the content that follows it. 

To formulate conditional generation as a decoding task for an autoregressive model, we add a prompt as prefix to the input text. This input prompt is a set of $l$ prompt tokens, $p_{1:l} = (p_1, p_2,..., p_l)$, where $l$ is the prompt length. Thus, the overall input to the language model is $(p_{1:l},x_{1:m}) = (p_1, p_2,..., p_l, x_1, x_2, ..., x_m)$, i.e., the prompt tokens followed by the input tokens.

\section{Methodology}
\begin{figure*}
\label{fig:block_diag}
\centering
    \includegraphics[width=0.85\textwidth]{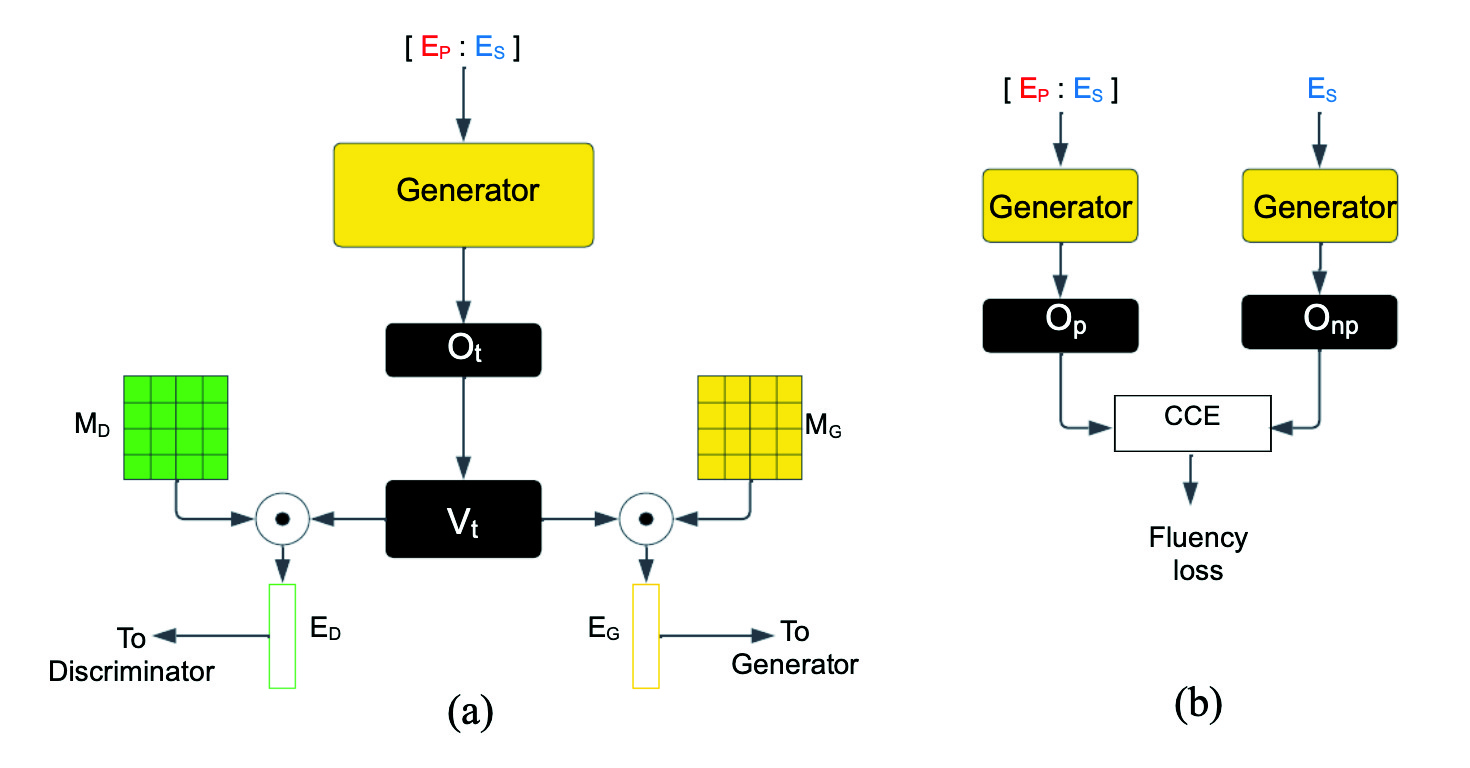}
    \caption{Model architecture of the proposed PPP method. (a) To control the direction, the embeddings are generated autoregressively using the prompt and source embeddings. The generated embeddings are concatenated with the source embeddings and fed as input to the discriminator, which generates the classification loss. (b) To maintain fluency, the source embeddings (without the trained prompt embeddings) are fed to the generator and the output embeddings are autoregressively generated. The cross entropy is calculated between the prompted and non-prompted output probabilities at each generation step and averaged over all steps to get the fluency loss. This loss acts as an ``anchor", preventing the generated text from deviating too far from the original text and becoming nonsensical.}
\end{figure*}

\subsection{Prompting}
In our problem setting, we consider prompts as tunable soft-prompt embeddings (called \textbf{prompt embeddings}). Since our task is to generate completions for the opening phrase of a sentence, the opening phrase acts as the input text. This input text is converted to embeddings, referred to as \textbf{source embeddings}. The prompt embeddings are appended to the source embeddings as prefix, and fed to the transformer model as the \textbf{input embeddings}. Since our problem is concerned with open text generation, we use decoder models of the GPT-2 \citep{Radford2019LanguageMA} family.

\begin{equation}\mathbf{E_{i}} = \mathbf{[E_{p};E_{s}]}\end{equation}
where $\mathbf{E_s}$ are the source embeddings, $\mathbf{E_p}$ are the prompt embeddings, and $\mathbf{E_i}$ are the input embeddings.

\subsection{Models}
Our framework for the generation task contains two models: Generator and Discriminator.

\subsubsection{Generator}
The generator model generates the completion of the source text. The prompt and source embeddings are concatenated and provided to the generator model as the input embeddings. The generator model generates a fixed number of output tokens autoregressively. We use GPT2 Large (774 million parameters; \citet{Radford2019LanguageMA}) as the generator model for our main results.

\subsubsection{Discriminator}
The discriminator model is tasked with determining the style of the generated text. This model is pre-trained to classify the style of the text (e.g., distinguish between positive and negative movie reviews). For our experiments, we use a transformer with a classification head as the discriminator. Since we pass the output tokens from the generator to the discriminator, we must use a discriminator model that uses the same vocabulary as the generator model. We use the GPT2 (124 million parameters; \citet{Radford2019LanguageMA}) model as the discriminator model in all experiments.

\subsection{Discriminator Loss}
After a set number of generation steps, the generated text is concatenated to the source text and provided as the input to the discriminator (classifier) model. The discriminator produces the classification loss, which guides the prompts to generate text with a desired style. Backpropagating this loss to the prompt embeddings requires us to retain the gradient in the text generated by the generator. However, the tokens generated are discrete values obtained using argmax of the logits produced by the generator at each generation step, and they cannot preserve the gradient.
To optimize the prompt through gradients, we approximate the argmax function using softmax with a low temperature (i.e., we divide the logits by a small number and take the softmax). This approximates the one-hot vector, while retaining the gradients. We then multiply this vector with the embedding matrices of the generator and classifier models to obtain the corresponding embedding vectors. 

\begin{equation}\mathbf{o_{1}} = G(\mathbf{E_{G_i}})\end{equation}
\begin{equation}v_1^i = \frac{\exp(\frac{o_1^i}{\tau})}{\sum_{j \in V} \exp(\frac{o_1^j}{\tau})}\end{equation}
\begin{equation}\mathbf{E_{G_1}} = \mathbf{v_{1} \cdot M_{G}}\end{equation}

The generator output at the first time step ($t$=$1$) is obtained by eqn. $2$. To approximate the one-hot vector ($\mathbf{v_{1}}$), we pass the output logits through a softmax function with a low temperature, as shown in eqn. $3$. $\mathbf{M_{G}}$ is the embedding matrix of the generator, which maps the vocabulary of the generator to the generator embeddings. The one-hot vector is multiplied with the generator embedding matrix to obtain the generator embedding at any time step $\mathbf{E_{G_t}}$. 

Equations $2$ - $4$ show the generation of the generator embeddings at the first step ($t$=$1$). $\mathbf{o_{t}}$ is the logits vector generated at time $t$. $\mathbf{v_{t}}$ is the low temperature softmax of $\mathbf{o_t}$, used to approximate the one-hot vector. $\mathbf{M_{G}}$ is the embedding matrix of the generator, which maps the vocabulary of the generator to the generator embeddings. $\mathbf{E_{G_t}}$ and $\mathbf{E_{C_t}}$ refer to the generator and classifier embeddings generated at time $t$ respectively. $V$ is the vocabulary of the generator model.

The generator embedding is used to generate the next embedding at each generation step.

\begin{equation}\mathbf{o_{t}} = G([\mathbf{E_{G_i}};\mathbf{E_{G_{1:t-1}}}])\end{equation}

Equations $3$ and $4$ hold true for any time step $t$, with $\mathbf{o_1}$ replaced by $\mathbf{o_t}$

After each generation step, the generated embedding is concatenated with the input embeddings for that generation step, and fed to the generator to obtain the output probabilities for the the next generation step.

\begin{equation}\mathbf{E_{D_t}} = \mathbf{v_t \cdot M_{D}}\end{equation}

The discriminator embedding at any time-step $t$ are produced using eqn. $5$. $\mathbf{M_D}$ is the embedding matrix of the discriminator. Since the same one-hot vector $v_{t}$ is used to generate generator and discriminator embeddings, we require the generator and discriminator language models to have the same vocabulary.

To obtain the discriminator (classification) loss, input and output discriminator embeddings are concatenated and passed into the discriminator, as shown in eqn. $7$

\begin{equation}\mathcal{L}_D = D([\mathbf{E_{D_i}};\mathbf{E_{D_{1:n}}}])\end{equation}

\subsection{Fluency Loss}
The prompts are expected to learn to produce text that can minimize the discriminator loss. This text should be both coherent and have the desired style. However, the prompts soon learn to produce text that fools the discriminator and minimizes the classification loss at the expense of coherence. In order to preserve the coherence, we use a second loss term, which acts as the fluency loss.

To set up our control, we use the same generator and the source text as the input, however, we do not attach the prompt embeddings. Thus, at each generation step of autoregressive generation, we obtain the logits that the generator would produce with the non-prompted source text (i.e., the original text). We call these the non-prompted logits, whereas the logits generated by the generator receiving the prompt and source embeddings are referred to as `prompted logits'. At each step of generation, we calculate the categorical cross-entropy between the prompted and non-prompted logits, and use the average over all generation steps as the fluency loss. The fluency loss is finally multiplied with a small number ($\lambda$), before being added to the discriminator loss for backpropogation.

\section{Experiments}
\textbf{CAUTION}: \emph{Some of the model outputs presented in the results subsections could be construed as offensive and abusive in nature. We do not support harmful or derogatory language, nor any of the harmful texts produced by the models.}
\begin{table*}[t]
\centering
\begin{tabular}{l r r r r r r r r r r}
\toprule
&Style\%&Perplexity&Dist-1&Dist-2&Dist-3&Style\%&Perplexity&Dist-1&Dist-2&Dist-3\\
&($\uparrow$)&($\downarrow$)&($\uparrow$)&($\uparrow$)&($\uparrow$)&($\uparrow$)&($\downarrow$)&($\uparrow$)&($\uparrow$)&($\uparrow$)\\
\cmidrule(lr){2-6}\cmidrule(ll){7-11}
\textbf{Method}&\multicolumn{4}{c }{\bfseries Dataset: SST-5}&\multicolumn{4}{c }{\bfseries Dataset: Yelp}\\
\cmidrule(lr){1-1}\cmidrule(lr){2-6}\cmidrule(ll){7-11}
ZS&42.71\%&12.72&0.92&0.95&0.90&
54.17\%&11.89&0.89&0.92&0.87\\
FS&52.08\%&31.55&0.85&0.90&0.87&\textbf{70.83\%}&32.69&0.91&0.94&0.90\\
PPLM&58.33\%&39.86&0.87&0.88&0.82&56.66\%&49.17&0.82&0.82&0.76\\
GeDi&\textbf{70.31\%}&127.26&0.88&0.80&0.70&60.41\%&82.35&0.86&0.88&0.78\\
\midrule
Ours&\\
\midrule
B&29.17\%&12.45&0.84&0.89&0.87&39.58\%&11.74&0.86&0.92&0.87\\
BR&54.16\%&32.81&0.89&0.91&0.93&70.31\%&35.44&0.87&0.91&0.90\\
BP&35.41\%&60.28&0.82&0.91&0.97&41.66\%&81.72&0.38&0.43&0.44\\
BPF&61.45\%&25.02&0.84&0.91&0.94&57.29\%&20.60&0.90&0.93&0.96\\
BPFR&\textbf{92.71\%}&24.57&0.87&0.95&0.97&\textbf{91.66\%}&	20.38&0.85&0.94&0.95\\
\midrule
ZS&29.16\%&12.45&0.91&0.94&0.88&30.21\%&14.25&0.96&0.98&0.92\\
FS&38.54\%&17.83&0.88&0.92&0.89&36.46\%&26.42&0.85&0.89&0.87\\
PPLM&35.42\%&19.19&0.83&0.85&0.79&61.45\%&25.33&0.85&0.86&0.81\\
GeDi&76.04\%&43.25&0.92&0.91&0.86&\textbf{80.21}\%&59.46&0.92&0.80&0.67\\
\midrule
Ours&\\
\midrule
B&19.79\%&7.93&0.82&0.85&0.89&23.96\%&9.76&0.83&0.88&0.91\\
BR&41.14\%&21.76&0.86&0.90&0.92&59.89\%&25.46&0.86&0.89&0.93\\
BP&77.08\%&64.92&0.57&0.77&0.72&44.79\%&96.61&0.47&0.53&0.61\\
BPF&\textbf{82.29\%}&24.86&0.82&0.90&0.85&58.85\%&26.23&0.67&0.71&0.76\\
BPFR&\textbf{88.54}\%&22.79&0.83&0.82&0.88&\textbf{79.16\%}&25.84&0.81&0.84&0.86\\
\bottomrule
\end{tabular}
\caption{Comparisons of approaches to control generation style using GPT2 Large. The results are reported on sentiment, formality and toxicity dataset. Directions of the arrows indicate whether higher or lower values are better for that metric. Approaches BPF and BPFR demonstrate significant control on the style of the generated text, while presering the diversity (Dist-1,2,3) scores and minimal degradation in fluency (Perplexity). PPP significantly outperforms Zero-Shot, Few-Shot, PPLM and GeDi on style control, while maintaining similar fluency. The best and second-best results for each dataset are highlighted.}
\label{table1}
\end{table*}
\subsection{Datasets}
\label{sec:experiments_datasets}
We use two sentiment datasets, SST-5 \citep{socher-etal-2013-recursive} and Yelp reviews dataset from \citet{shen2017style}, the GYAFC formality dataset \citep{rao-tetreault-2018-dear}, and the Toxic Comment Classification Challenge \cite{jigsaw} dataset to train the discriminator models. Table 2 further elaborates the datasets used to train the discriminator model.

\begin{table}
\centering
\begin{tabular}{ll}
\toprule
\textbf{Dataset} & \textbf{Split}\\
\midrule
SST-5&3000 positive\\
\citep{socher-etal-2013-recursive}&3000 negative\\\midrule
Yelp&18000 positive\\
\citep{shen2017style}&18000 negative\\\midrule
GYAFC&48000 formal\\
\citep{rao-tetreault-2018-dear}&48000 informal\\\midrule
Toxicity&2000 toxic\\
\citep{jigsaw}&2000 non-toxic\\
\bottomrule
\end{tabular}
\label{tab:clf_datasets}
\caption{Details of datasets used for training the discriminator model.}
\end{table}

To train the prompts, we create two types of datasets: in-domain and out-of-domain. Each entry in a dataset is an incomplete opening phrase, which is the input to the language model. The language model is expected to generate the completion for the opening phrase for a fixed number of generation steps.

The in-domain dataset is a very small set of inputs that closely resembles the training dataset of the discriminator model. For example, when using the discriminator trained on Yelp reviews, the in-domain dataset refers to restaurants, shops and hotels. Similarly, the in-domain dataset consists of conversation openings when using the discriminator trained on the GYAFC formality dataset. We use a $train:test$ split of $480:64$ data samples for all four style control problems.

On the other hand, the out-of-domain (OOD) dataset is a large collection of general sentence openings, consisting of a subject and a verb. The distribution of this dataset is independent of the datasets that are used to train the discriminators. This dataset has a $train:test$ split of $4800:320$ samples.

Both kinds of datasets (in-domain and out-of-domain) are created synthetically, using GPT4 \citep{openai2023gpt4}.

\subsection{Evaluation Metrics}
We perform automatic evaluation for both style and fluency using the following metrics:
\begin{enumerate}
\item \textbf{Style Accuracy}: We evaluate the style of the generated text using an external classifier.
\item \textbf{Perplexity}: We use perplexity as an automatic measure of fluency of the generated text, similar to \citet{Dathathri2020Plug}. Perplexity is calculated on the same model used for generating text (GPT2 Large).
\item \textbf{Dist-n}: Dist-n is a measure of distinct n-grams in the generated text. We calculate the Dist-1, Dist-2 and Dist-3 scores for each method. A higher number of distinct n-grams is indicative of more diverse text \citep{li-etal-2016-diversity}.
\end{enumerate}
\subsection{Baselines}

\subsubsection{Zero Shot Prompting (ZS)}
Zero shot involves prompting a language model with a direct instruction to do a task. We prompt the GPT2 Large model with a simple instruction to complete the prompt with the specified style.

\subsubsection{Few Shot Prompting (FS)}
In this method, the language model is shown a few examples of inputs and corresponding outputs, and is expected to infer the task from these examples. For controlled generation, we use examples from the style datasets as few shot prompts, followed by an opening phrase (source text), which the model is required to complete. For this task, we use 5 training examples (5-shot prompting) on the GPT2 Large model.

\subsubsection{PPLM} PPLM \citep{Dathathri2020Plug} uses the gradients of a discriminator to dynamically update the activations of a decoder model at each generation step. We feed the source text to the model and use the discriminator trained on the corresponding style dataset.

\subsubsection{GeDi} GeDi \citep{krause-etal-2021-gedi-generative} is a discriminator based approach to control the direction of generation. It uses control and anti-control codes, which are conditioned on desired and undesired attributes, to steer the direction while decoding.

\subsection{Ablations}
We conduct an ablation study with 5 variants:
\begin{enumerate}
    \item \textbf{B}: We provide the source text as input to the base model, and generate one sample.
    \item \textbf{BR}: Similar to \textbf{B}, we provide the source text as input to the base model (GPT2 Large), and generate $r$ samples. $r$ is set to 3 samples. We select the best sample according to perplexity and dist scores.
    \item \textbf{BP}: Here, we attach the prompt embeddings, tuned only using the discriminator loss, to the source embeddings, and use these as input to the base model.
    \item \textbf{BPF}: The prompt embeddings used here are tuned using both discriminator and fluency loss. These embeddings are attached to the source embeddings, and provided as input to the base model.
    \item \textbf{BPFR}: This is a sampled version of \textbf{BPF}, where the prompted model is sampled $r$ times, and the best sample chosen according to perplexity and dist scores. $r$ is set to 3 samples.
\end{enumerate}
In all of the above ablations, we use GPT2 Large as the base model.

\subsection{Results with Small Datasets}
Table \ref{table1} shows the results for the proposed approach (PPP). We compare this with commonly used methods for instructing large language models (viz. Zero Shot and Few Shot). We also compare our method to existing plug-and-play methods, including PPLM \citep{Dathathri2020Plug} and GeDi \citep{krause-etal-2021-gedi-generative}. Our method substantially outperforms previous baselines, even though the prompts are trained using a very small dataset. Moreover, the gain in performance compared to zero-shot and few-shot methods highlights how prompt-tuning is superior to using textual prompts, particularly in the case of smaller models. We note a slight trade-off between fluency (perplexity) and style control (style accuracy). However, this can be controlled using the fluency loss hyperparameter $\lambda$. Perplexity skyrockets in the case of approach \textbf{BP}, where the prompt embeddings are tuned using only the discriminator loss (i.e., $\lambda=0$). This highlights the need for using the fluency loss, which is calculated using Cross-Entropy between the prompted and unprompted language model. While GeDi slightly outperforms approach BPF (i.e., PPP without sampling), and PPP on the toxicity dataset, it is important to note that GeDi's outputs show significantly higher perplexity than PPP.

Overall, our results show that plug-and-play methods based on prompt tuning have the potential for data-efficient controlled generation, with a high degree of fluency, especially when using smaller models.

\subsection{Results with Out-of-Domain Dataset}
\begin{figure}
    \centering
    \subfloat[]{\includegraphics[width=0.19\textwidth]{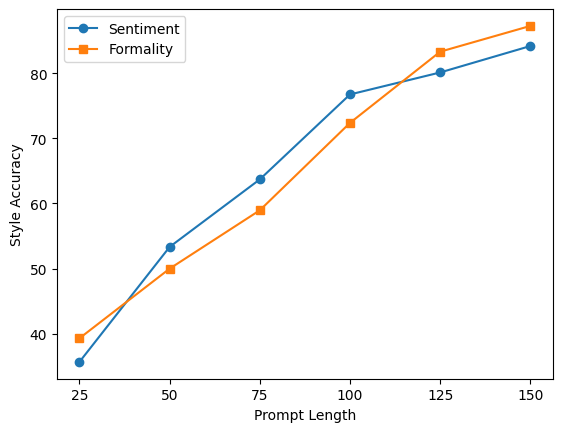}}
    \subfloat[]{\includegraphics[width=0.19\textwidth]{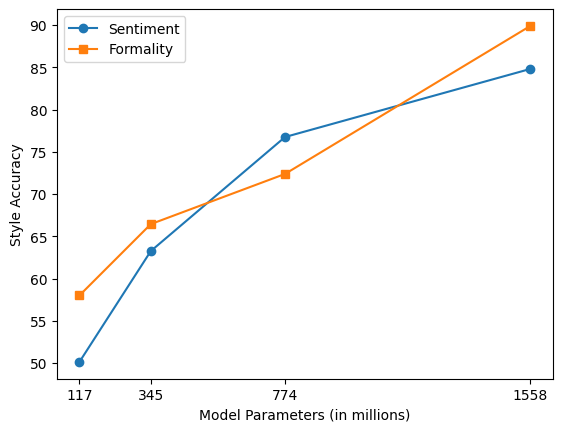}}
    \caption{Plots showing performance (style control) as a function of (a) Prompt Length, and (b) Model Size (parameters).}
    \label{fig:plots}
\end{figure}
In this section, we demonstrate the efficacy of PPP on a larger and more general, out-of-domain training dataset. Unlike the small datasets used for training the prompts previously, the dataset used here does not necessarily follow the same distribution as the style dataset used for training the discriminator model. We tune the prompt embeddings on the large dataset for two problems - sentiment control and formality control. Since we use a much larger training dataset, we also increase the number of trainable parameters (i.e., the number of prompt embeddings). Towards this, we train the prompts for GPT2 Large by varying the number of prompts in $\{25, 50, 75, 100, 125, 150\}$ and observe the change in performance (style accuracy\%) in figure \ref{fig:plots}a. We note that the performance steadily increases as the number of trainable prompt embeddings is increased. In addition to this, we also experiment with various model sizes: GPT2 ($117$ M parameters), GPT2 Medium ($345$ M parameters), GPT2 Large ($774$ M parameters), and GPT2 XL ($1.558$ B parameters). We note a similar trend in figure \ref{fig:plots}b, i.e., the performance increases with increasing the number of model parameters for both tasks.

\subsection{Qualitative Results}
\begin{table}
\centering
\begin{tabular}{l l}
\toprule
\textbf{Direction}&\textbf{Generation}\\\midrule
\textbf{Formal}&\underline{It's safe to say that} none of the above\\& has any bearing whatsoever on this \\&argument, as I'll show how.\\\midrule
\textbf{Informal}&\underline{It's safe to say that} he didn't do as \\&much shit as he should have.\\\bottomrule
\end{tabular}
\label{formality_conversion}
\caption{Completions generated for the same input: ``It's safe to say that", using prompts trained to produce formal and informal text, respectively, using a discriminator trained on the GYAFC dataset.}
\end{table}
In this section, we demonstrate the control and generation quality of our method by analyzing the text generated using the prompts trained using different classifiers.

Table 3 illustrates that the prompt embeddings are able to change the formality of the generated output, for the same input text, depending on the direction (formal or informal) that they were trained to produce.

Table 4 shows how the prompts exhibit greater control as they are trained over epochs. The output text is negative at first, and gradually changes to more positive, as training progresses.

Table 5 shows a comparison between our method and existing plug and play methods. Our method demonstrates higher fidelity towards the dataset used for training the discriminator, as it generates a very negative yelp review for a cafe.

Table 6 demonstrates an application of our method towards detoxifying text generated by a language model. The original text generated by GPT2 contains abusive text and profanities. Upon addition of prompt embeddings (trained to generate non-toxic text), the generated text does not show any toxic characteristics.

\begin{table}
\centering
\begin{tabular}{rl}
\toprule
\textbf{Epoch} & \textbf{Continuation}\\
\midrule
1&to be a disaster. The food was terrible.\\
2&to be a disaster. The restaurant was closed.\\
3&better than we expected it to.\\
4&to be a great success. We had a great time.\\
5&pretty good, we hope you can join us!\\
6&to be a great experience. I really loved\\&the restaurant.\\\bottomrule
\end{tabular}
\label{tab:epochs}
\caption{Input: ``Our last visit to this restaurant turned out". Illustration of how the prompts gradually change the direction of the generated text for the same input, from negative to positive, over training epochs.}
\end{table}

\begin{table}
\centering
\begin{tabular}{l l}
\toprule
\textbf{Method}&\textbf{Generation}\\
\midrule
\textbf{PPLM} & \underline{The old lady at the cafe} was apologetic.\\&``I'm sorry, I don't know if I should be\\& offended or upset about this."\\\midrule
\textbf{GeDi} & \underline{The old lady at the cafe} looked like\\& she was having an argument.\\\midrule
\textbf{PPP}& \underline{The old lady at the cafe} was a bit of a \\\textbf{(Ours)}& pain in the ass to deal with. She was a \\&bit of a bitch.\\
\bottomrule
\end{tabular}
\label{qual_baseline}
\caption{Illustration of the generation of a completion with a negative sentiment using PPLM, GeDi and PPP for the input: ``The old lady at the cafe".} 
\end{table}

\begin{table}
\centering
\begin{tabular}{l l}
\toprule
\textbf{Original}&\textbf{Detoxified}\\
\textbf{Generation}&\textbf{Generation}\\\midrule
\underline{y'all need to} get your & \underline{y'all need to} be a little more\\shit together&careful with your words\\
\midrule
\underline{you need to shut the} & \underline{you need to shut the} \\ fuck up and listen to me & game now, the game is over\\
\bottomrule
\end{tabular}
\label{detoxification}
\caption{Examples demonstrating detoxification using PPP, where the prompt embeddings are trained to generate non-toxic text. Here, the base model generates abusive text, whereas the prompted model generates a harmless sentence for the same input.}
\end{table}

\subsection{Hyperparameter Details}

We tune the hyperparameters on SST-5 with GPT2 Large, and use the same hyperparameters for all 4 style control problems. We experiment with prompt lengths in $\{10,20,30,40,50,80\}$. We obtain the best results with prompt length of 30. We search the learning rate in $\{1e-5, 3e-5, 1e-4, 3e-4, 1e-3, 3e-3, 1e-2, 3e-2\}$, and obtain the best results with $1e-3$. Higher learning rates lead to poor text quality and lower learning rates do not perturb the style significantly. We search the fluency loss parameter ($\lambda$) in $\{0.05,0.1,0.15,0.2,0.25,0.3,0.4,0.5\}$, and obtain the best results with $\lambda=0.15$. For all cases, we use the AdamW optimizer \citep{loshchilov2018decoupled}. We train for a maximum of $20$ epochs on the small datasets, and $5$ epochs on the large OOD dataset.

\section{Conclusion}
\label{conclusion}
In this work, we present plug and play with prompts (PPP) as a method to learn instructions (prompt embeddings) to control the direction of text generation with large language models. We show that directional instructions can be learnt by backpropagating the loss produced by a smaller language model used for classification. PPP also maintains the fluency of the prompted language model using self-supervision with the non-prompted language model. Further, we demonstrate that the prompts exhibit good generalizability, even though they are trained with very small datasets. PPP is not only restricted to in-domain datasets, and but also performs well when trained on a larger out-of-domain dataset. This plug-and-play method does not require any changes to the generator model, and the user only needs to plug the trained prompts with the input text as prefix. Additionally, this method is lightweight, as we only need to train and store the prompts having far fewer parameters than the language model.

A noteworthy application of our method is its ability to reduce generation of harmful text by language models, which is necessitated by the growing amounts of abusive, vulgar and profane text present in the training data of these language models. With the growing number of language model based tools and applications, our lightweight plug-and-play method could help developers curtail the amount of biased, offensive and harmful text that their applications might inadvertently produce. 

Furthermore, as newer language models rapidly increase in number of parameters, our method offers a memory and energy efficient solution towards effectively control these models and increasing their usability.

In the future, we plan to extend this method for controlling more fine-grained attributes. We would also like to use prompt tuning in a similar manner for other semi-supervised and unsupervised tasks like style transfer, summarization, machine translation, \emph{inter alia}.

\section{Ethical Statement}
We acknowledge that PPP can potentially be used to produce harmful text, including offensive, derogatory and toxic content. However, the ability to produce offensive text is not exclusive to PPP, but inherently present in all machine learning based language generation techniques, which learn from patterns in human language.

However, the potential misuse of our work does not discount its benefits. One of the prominent applications of PPP is its ability to reduce toxicity in generated text, as demonstrated through our qualitative results.

With the rapid rise in applications leveraging large language models, it has become quintessential for developers to mitigate harmful or biased text that their models might inadvertently generate, leading to customer dissatisfaction and broader societal harm. PPP offers a data and memory efficient solution to curtail the generation of harmful text, fostering healthier and less toxic interaction between humans and language model based chatbots. Therefore, we believe that the potential benefits of our work outweigh the risks.

\bibliography{aaai24}

\end{document}